\documentclass[journal ]{new-aiaa}
\usepackage[utf8]{inputenc}
\usepackage{textcomp}
\usepackage{subcaption}
\usepackage{caption}
\usepackage{graphicx}
\usepackage{epstopdf}
\usepackage[ruled,vlined]{algorithm2e}
\usepackage{amsmath}
\usepackage[version=4]{mhchem}
\usepackage{siunitx}
\usepackage{longtable,tabularx}
\usepackage{epstopdf}

\title{Modular Multi Target Tracking Using LSTM Networks}

\author{Rishabh Verma, R Rajesh and MS Easwaran}
\affil{Centre for Airborne Systems,DRDO

 Bengaluru,India -560037}

\begin{document}

\maketitle

\begin{abstract}
The process of association and tracking of sensor detections is a key element in providing situational awareness. When the targets in the scenario are dense and exhibit high maneuverability, Multi-Target Tracking (MTT) becomes a challenging task. The conventional techniques to solve such NP-hard combinatorial optimization problem involves multiple complex models and requires tedious tuning of parameters, failing to provide an acceptable performance within the computational constraints. This paper proposes a model free end-to-end approach for airborne target tracking system using sensor measurements, integrating all the key elements of multi target tracking - association, prediction and filtering using deep learning with memory. The challenging task of association is performed using the Bi-Directional Long short-term memory (LSTM) whereas filtering and prediction are done using LSTM models. The proposed modular blocks can be independently trained and used in multitude of tracking applications including non co-operative (e.g., radar) and co-operative sensors (e.g., AIS, IFF, ADS-B). Such modular blocks also enhances the interpretability of the deep learning application. It is shown that  performance of the proposed technique outperformes conventional state of the art technique Joint Probabilistic Data Association with Interacting Multiple Model (JPDA-IMM) filter.
\end{abstract}

\section{Introduction}
\lettrine{M}{ulti-Target} tracking (MTT) in a clutter environment with noisy measurements is a challenging task. A reliable tracking system should be able to accurately estimate the state of the targets, which is often difficult due to the association capabilities and measurement imperfections. Kalman filter (KF) \cite{10.1115/1.3662552} has been a widely used technique for MTT, and many of its variants such as unscented KF (UKF) \cite{Julier2000ANM}, extended KF (EKF) \cite{10.1117/12.280797} and particle filter (PF)\cite{978374} further improve the tracking accuracy. These techniques require pre-defined mathematical motion models of the targets for tracking and are based on Bayesian tracking theory. However, it is highly unlikely to have precise mathematical models of the targets in advance especially in  military scenarios. Since, the performance of such techniques rely on the accuracy of the models, they suffer a serious degradation in the performance due to model imperfections.  To improve performance Multiple Model (MM) \cite{489270} algorithm uses more than one  filters simultaneously to track the target. Further improvements were made in Interactive Multiple Model (IMM) \cite{1299} where the weights of the models are derived according to the changing observations.

A typical MTT process involves association of measurements to tracks, managing birth \&  death of tracks, prediction of track parameters and filtering of measurements. The process of association in a clutter environment is a NP- hard problem. Many sophisticated algorithms have been developed in the past including the multiple hypothesis tracker (MHT) \cite{1102177} which generates a tree of potential hypothesis for each target, therefore leads to the combinatorial explosion and computational overload with increase in number of targets. Joint Probabilistic Data Association (JPDA) \cite{1145560} computes all possible joint events for calculating the association probabilities and requires knowledge of clutter rate and detection probability for computations.

Recently, deep learning based techniques have shown promising results in solving the problem of tracking and association especially in pedestrian tracking \cite{9102884,9011317,BABAEE201969} from videos. Anton Milan et al. \cite{10.5555/3298023.3298181} used Recurrent Neural Network (RNN) based model for prediction, birth/ death and LSTM based model for association in a two-dimensional (2D) scenario and showed that the neural network without any prior knowledge about target dynamics, clutter distributions etc, is able to perform prediction, data association, and filtering in the context of video tracking. For radar tracking, C.Gao et al. \cite{8557284} showed that multi layer RNN structures gives better filter results as compared to single layer while both the model outperformed the IMM. Jingxian Liu et al. \cite{LiuWX20} used Bi-directional LSTM based filtering algorithm in a 2D clutter free environment. Huajun Liu et al. \cite{DeepDA} showed LSTM based models performs better association as compared to classic models like JPDA and Hungarian algorithm (HA) in a 2D clutter environment. The papers above only consider parts of MTT problem in the context of sensor measurements and contain targets with low agility and range.

This paper makes the following contributions:
\begin{enumerate}
  \item It considers a 3D scenario with clutter and high maneuvering targets as compared to most of the previous work, that are focused on 2D models and/or clutter free environment.
  \item It provides an end-to-end model based on deep neural networks for multi-target tracking, consisting- filtering, prediction, track management and association with modular building blocks.
  \item It uses a Bi-directional LSTM based model for association process and shows that it outperforms LSTM based model in efficient training and performance.
  \item It provides modular deep learning blocks which can be adopted for MTT problems for both co-operative and non co-operative sensors.
  \item It characterizes the performance of the end to end approach with conventional approach using Generalized Optimal Sub-Pattern Assignment (GOSPA) \cite{Rahmathullah_2017} metric, showing the effectiveness of the proposed modular blocks.
\end{enumerate}

\section{Multi Target Model}
\label{Section:MTT}
The scenario under consideration has multiple and dense high maneuvering airborne targets. The scenario is sensed by a sensor which induces imperfections modeled as measurement error and false detections originating from clutter. Such a scenario is typical of a large scale military combat.

\subsection{Motion Model}
This paper uses a particle model \cite{motion_model} with three degree of freedom for the aircraft, with the assumption of no slide slip, no angle of attack and no effect of rotation of the earth. The motion of the aircraft with ground coordinate system as inertial frame can be expressed as Eq.~\eqref{motion:kinetic}, where $x$, $y$, $z$ are the position coordinates of the aircraft, $v$ is linear velocity, $\phi_{\textrm {p}}$ is pitch angle and $\phi_{\textrm {a}}$ is azimuth angle of the aircraft. The six dimensional state vector is defined as $[x,y,z,v,\phi_{p},\phi_{a}]$.
\begin{equation}
  \label{motion:kinetic}
  \begin{cases}
     \dot {x}=v \cos \phi_{\textrm {p}} \cos \phi_{\textrm {a}} \\
     \dot {y}=v \cos \phi_{\textrm {p}} \sin \phi_{\textrm {a}} \\
     \dot {z}=v \sin \phi_{\textrm {p}} \\
  \end{cases}
\end{equation}
The simplified dynamic equations can be described as Eq.~\eqref{motion:simplified}, where $n_{x}$ is normal overload and $n_{z}$ is tangential overload. The motion of the aircraft is controlled by the control variables $[\phi_{r},n_{z},n_{x}]$. The truths are generated by randomly changing the control variables.
\begin{equation}
  \label{motion:simplified}
  \begin{cases}
    \dot {v} =g(n_{x} -\sin \phi_{\textrm {p}}) \\
     \dot {\phi_{\textrm {p}}} =\frac {g}{v}(n_{z} \cos \phi_{\textrm {r}} -\cos \phi_{\textrm {p}}) \\
     \dot {\phi_{\textrm {a}}} =\frac {g}{v \cos \phi_{\textrm {p}}}n_{z} \sin \phi_{\textrm {r}} \\
  \end{cases}
\end{equation}
\subsection{Measurement Model}
The measurement model for each measurement the sensor (e.g., radar) outputs matrix $M=[r,\phi^r_{p},\phi^r_{a},v_{r}]$, where $r$ is radial distance of target, $\phi^r_{p}$, $\phi^r_{a}$ are pitch and azimuth angles of target  and $v_{r}$ is radial velocity of the target $w.r.t.$ the radar. The four dimensional measurement vector for the sensor is calculated using Eq.~\eqref{motion:measurement}.
\begin{equation}
  \label{motion:measurement}
M = \begin{bmatrix}
\sqrt{x^2 + y^2 + z^2} \cr
tan^-1(z/\sqrt{x^2+y^2}) \cr
tan^-1(y/x) \cr
\frac{v \cdot r}{|r|}
\end{bmatrix} +N
\end{equation}
Where $ N \sim \mathcal{N}(\mu,\sigma^2) $ is measurement noise added with co-variance matrix  $\sigma^2 = diag([\sigma^2_d,\sigma^2_{\phi_{p}},\sigma^2_{\phi_{a}},\sigma^2_v])$, where $\sigma^2_d,\sigma^2_{\phi_{p}},\sigma^2_{\phi_{a}}$ and $\sigma^2_v$
are variance of radial distance, pitch angle, azimuth angle and radial velocity respectively of the target w.r.t. the radar.
The above models are used for the generation of truth and measurements for training deep neural networks.
\begin{figure}[ht]
\centering
\includegraphics[scale=1]{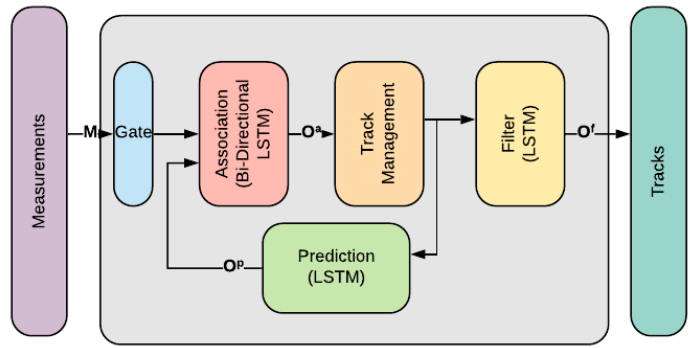}
\caption{Multi Target Tracker}
\label{fig:multi-target-tracker}
\end{figure}
\section{Deep Learning blocks for Multi-target tracking}
The building blocks of the multi-target tracker is shown in Fig.~\ref{fig:multi-target-tracker}. The main blocks are association block which associates a measurement to a track, track management block which takes care of birth and death of the tracks, prediction block which predict the state of the established tracks and filter block which refines the track positions. A gate is used before association block in case when the number of measurements exceed the capacity of association module, it break the measurements into small clusters before feeding to the association module. The gate helps on scaling the algorithm to large number of targets.

The process of prediction and filtering requires the knowledge of both the current and previous states of the aircraft. This task is realized through Long Short-Term Memory (LSTM) \cite{cho-etal-2014-learning}, which solves the vanishing and exploding gradients problem of RNN using multiple gates. These networks inherently have a memory element which helps in predicting any time series.

Association of a measurement to the correct track requires the knowledge of other tracks and measurements at that instance. This requires a memory element with Bidirectional information flow. Hence, Bi-directional LSTM \cite{650093} which initializes two independent LSTMs are used together running in opposite directions capturing both backward and forward information. Track Management implements a birth-death process is implemented using a rate based algorithm.

The measurement matrix $M$ of dimension $n \times 4$ where $n$ is number of measurements, along with matrix $O^p$ which is output from prediction block and has dimension $m \times 6$ where $m$ is the number of established tracks are fed to association block. Association module either associates each track to one of the measurements or NONE, and outputs matrix $O^a$ with dimension $(n+1)\times m$.  The unassociated tracks and measurements go through the track management module, where either the new tracks are formed or the idle tracks are terminated. The filter module refines the measurements for each established tracks and outputs the position, meanwhile the prediction module use measurements to predict the next states of each track and whole cycle repeats.

It can be seen that proposed design is modular and can be used for multiple cooperative and non cooperative sensors. These blocks can be independently tested, integrated together or in parts depending on the sensor of interest.
\begin{figure}[ht]
  \begin{subfigure}{.50\textwidth}
\centering
\includegraphics[scale = 1]{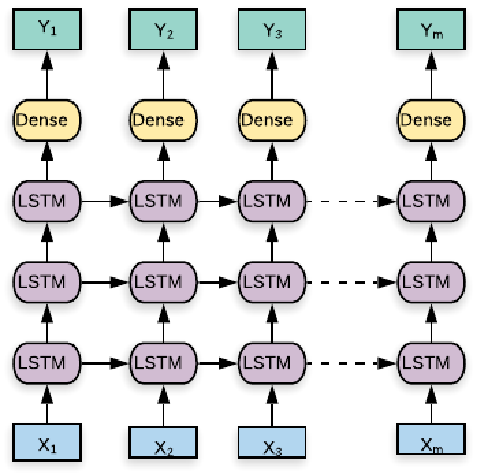}
\caption{LSTM Model}
\label{LSTM}
\end{subfigure}
\begin{subfigure}{.5\textwidth}
  \includegraphics[scale = 1]{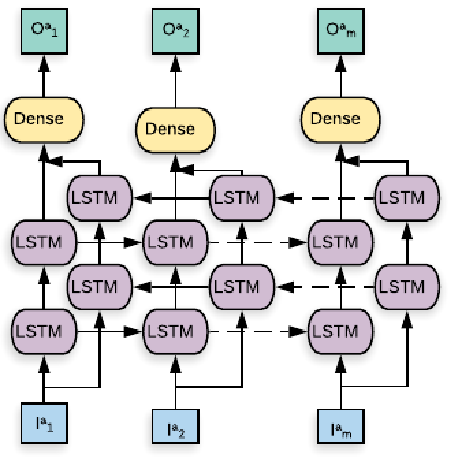}
  \caption{Bi LSTM Model}
  \label{bilstm}
\end{subfigure}
\caption{Deep Learning Models}
\end{figure}

\section{Model Free Tracker}
\subsection{Data Generation For Training}
Random training tracks are generated with linear velocity $v$ of targets ranging $[100,600]$ m/sec, linear acceleration varying from $[-2g,2g]$ and normal acceleration ranging from $[-2g,9g]$ according to motion model Eq.~\eqref{motion:kinetic}. Tracks are simulated at interval of $0.1$ sec. Radar measurements are generated by sampling the tracks at the interval of $5$ sec and adding measurement noise with variance matrix $\sigma^2=diag([\sigma^2_d,\sigma^2_{\phi_{p}},\sigma^2_{\phi_{a}},\sigma^2_v])$, where $\sigma^2_d = 30m$, $\sigma^2_{\phi_{p}}$ and $\sigma^2_{\phi_{a}} = 0.5^\circ$ and $\sigma^2_v = 5m/s$. $50,0000$ random tracks are generated with $100$ measurements each, and are divided in training and validation set in 80:20 ratio.

\subsection{Association}
\begin{figure}[ht]
\begin{subfigure}{.5\textwidth}
\centering
\includegraphics[scale = 1]{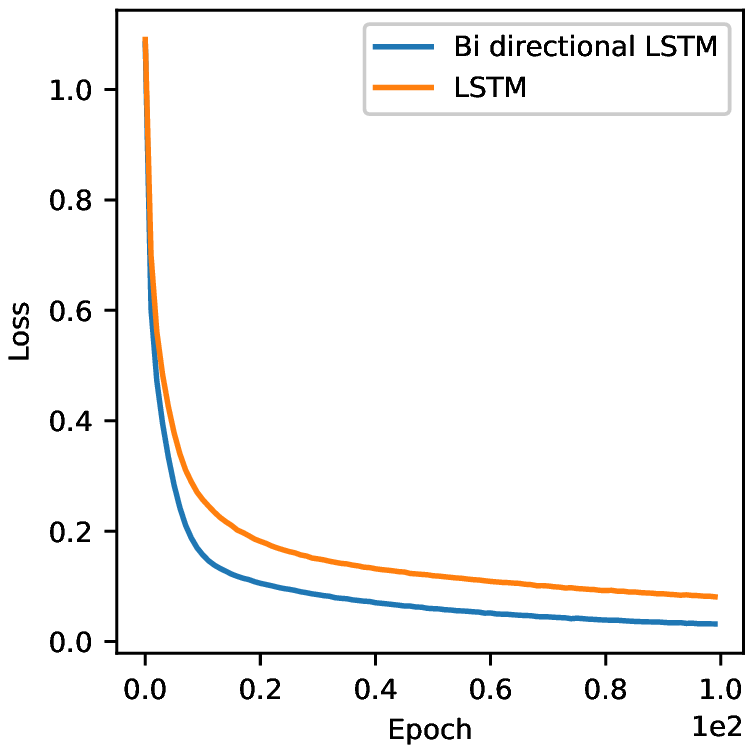}
\caption{Loss vs Epoch}
\label{bi:train1}
\end{subfigure}
\begin{subfigure}{.5\textwidth}
  \includegraphics[scale = 1]{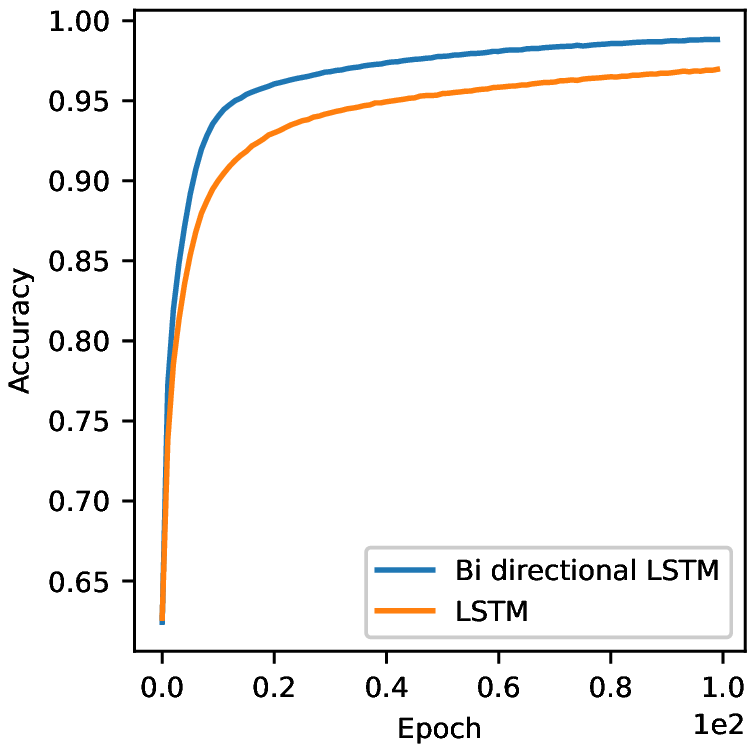}
  \caption{Accuracy vs Epoch}
  \label{bi:train2}
\end{subfigure}
\caption{Association Model}
\label{association}
\end{figure}

\subsubsection{Model}
Two different models are trained with the same set of training data and compared on the metric of accuracy and training convergence. A two layer Bidirectional LSTM model with hidden layer nodes = $128$, and a two layer LSTM model with hidden layer units = $256$ as described in Fig.~\ref{bilstm} are compared. Loss function used is categorical crossentropy, where for each established track the assignment loss is calculated using Eq.~\eqref{bilstm:Loss}, where $t_{i}$ is the ground truth  and $s_{i}$ is the predicted assignment score for the measurement $i$ to the current track from the sigmiod layer and $n$ is number of measurements.
\begin{equation}
  \label{bilstm:Loss}
    Loss = -\sum_{i}^{n}t_{i} log (s_{i})
\end{equation}
\subsubsection{Input}
The association module at each time step gets input from both radar as matrix $M$ after gating and prediction module as matrix $O^p$, where $M_{l}=[r,\phi^r_{p},\phi^r_{a},v_{r}]_l$, $l \in[1,n]$, where $n$ is the total number of measurements for current time $t$. And $O^p_k = [x,y,z,v_x,v_y,v_z]_{k}$, $k\in[1,m]$ where $m$ is number of established tracks. $I^a$ is $n$ x $m$ dimensional matrix, where $I^a_k$ is $k^{th}$ input sequence to the model.
$$
I^a_{lk}= abs\begin{pmatrix}\begin{bmatrix}
r*cos(\phi^r_{p})*cos(\phi^r_{a}) \cr
r*cos(\phi^r_{p})*sin(\phi^r_{a})\cr
r*sin(\phi^r_{p}) \cr
v_r*cos(\phi^r_{p})*cos(\phi^r_{a}) \cr
v_r*cos(\phi^r_{p})*sin(\phi^r_{a})\cr
v_r*sin(\phi^r_{p}) \cr
\end{bmatrix}_{l}-O^p_k\end{pmatrix}
$$
\subsubsection{Training} Model is trained for 100 epochs where each epochs contain $10,000$ episodes. The total number of dense tracks are chosen randomly between 0 to 10. Model is trained with Adam optimizer\cite{DBLP:journals/corr/KingmaB14} with learning rate of $0.001$. It can be seen from Fig.~\ref{association} that the Bi-directional LSTM outperformes the LSTM in the loss and accuracy metrics with $58.66\%$ and $2.02\%$ respectively.
\subsubsection{Output}
The output matrix $O^a$ of the model is an association matrix, where $O^a_{lk}$ $\in[0,1]$ and denotes the likelihood of measurement $l$ being associated to current track $k$, for each track $k$ $argmax$ is applied to find the associated measurement $l$, $l\in[0,n]$, where $0$ means unassociated and $1,2..n$ are indices of the measurements from the radar.
\begin{figure}[ht]
  \centering
\includegraphics[scale = 1]{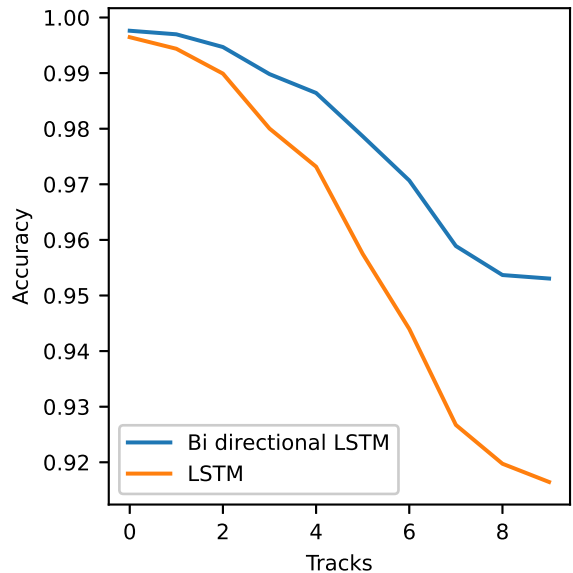}
\caption{Performance on test data}
\label{bi:acc}
\end{figure}
\subsubsection{Performance}
The model is tested on the different scenarios containing crossing tracks ranging from 1 to 8 with motion and measurement model as described in ~Section \ref{Section:MTT}. The measurements also contain false detections. As expected it can be seen in Fig.~\ref{bi:acc} that the Bi directional LSTM model shows better performance in associating the measurements to the tracks, especially when the number of tracks are higher. Also when the tracks are increased from 1 to 10 there is only $4.41\%$ reduction in accuracy of association.

\subsection{Prediction Module}
\begin{figure}[ht]
  \begin{subfigure}{.50\textwidth}
\centering
\includegraphics[scale = 1]{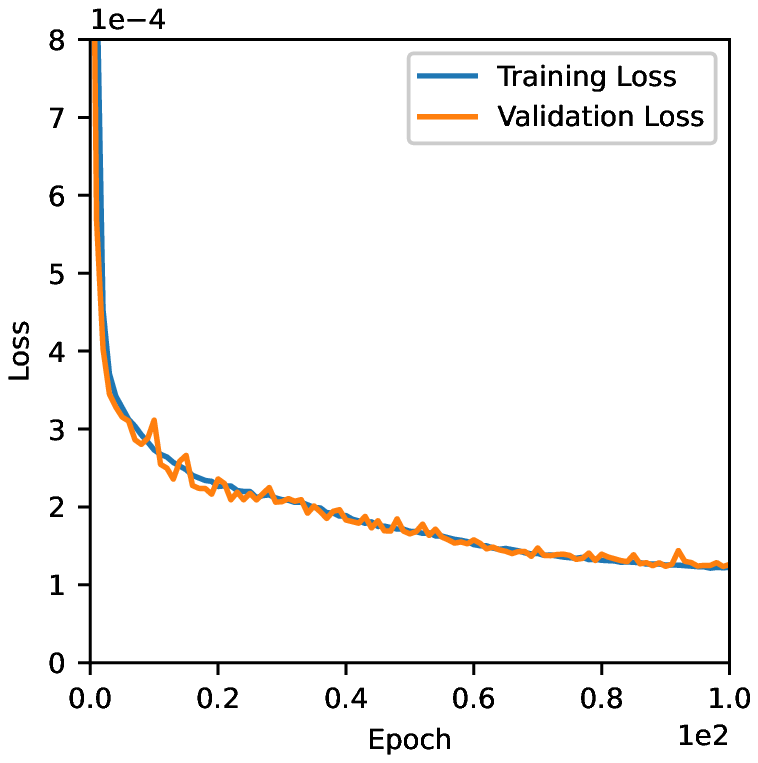}
\caption{Position Prediction}
\label{pre_p:train1}
\end{subfigure}
\begin{subfigure}{.50\textwidth}
  \includegraphics[scale = 1]{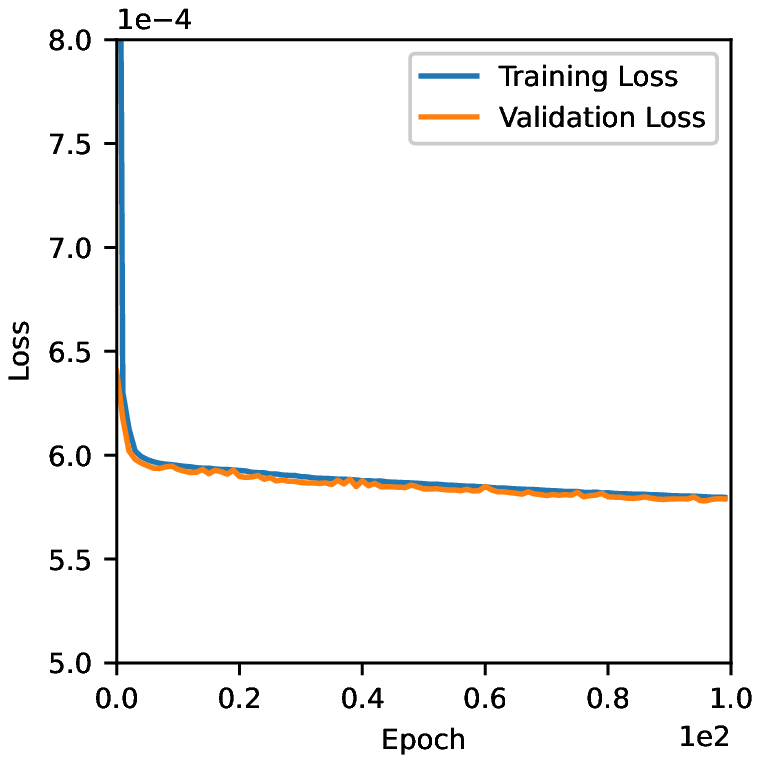}
  \caption{Velocity Prediction}
  \label{pre_v:train1}
\end{subfigure}
\caption{Prediction Model}
\label{Prediction}
\end{figure}
\subsubsection{Model}
For prediction, two different LSTM based models are used, one for predicting the position coordinates for time  $t+1$, and the other for the predicting the radial velocity components along the coordinate axis for time $t+1$ from the measurement state at time $t$. Both models are identical in shape and training hyper-parameters. A 3 layers stacked LSTM model as described in Fig.~\ref{LSTM} with number of nodes in hidden layer = 512 for all layers. Loss function used is Mean Squared Error (MSE), where loss is calculate as Eq.~\eqref{mse}, where $n$ is number of states, $y_{i}$ is ground truth of $i^{th}$ state and $x_{i}$ is predicted value.
\begin{equation}
  \label{mse}
    Loss = \frac{1}{n}\sum_{i=1}^{n}(y_{i} - x_{i})^{2}.\\
\end{equation}
\subsubsection{Input}
The input $I^p$ to the model are the current measurements of the tracks. $$I^p_k = [x,y,z,v_x,v_y,v_z]_{k }$$ where $I^p_k$is the input for track $k\in[1,m]$ where $ x,y,z $ are position coordinates and $ v_x, v_y, v_z $ are components of radial velocity $w.r.t.$ to coordinate axis and $m$ is number of current tracks.
\subsubsection{Training} Model is trained on $40,000$ tracks with 100 measurements each $5$ secs apart with Adam optimizer with learning rate of $0.001$. From Fig.~\ref{Prediction} it can be seen that the network is not over-fitting and provides good validation loss for both position and velocity prediction.
\subsubsection{Output}
The output matrix $O^p_k$ of the model is input for the association module. $$O^p_k = [x,y,z,v_x,v_y,v_z]_{k}$$ $k\in[1,m]$ where $m$ is number of current tracks, $ x,y,z $ are predicted position coordinates and $ v_x, v_y, v_z $ are components of radial velocity $w.r.t.$ to coordinate axis.

\begin{figure}[ht]
  \begin{subfigure}{.5\textwidth}
\centering
\includegraphics[scale=1]{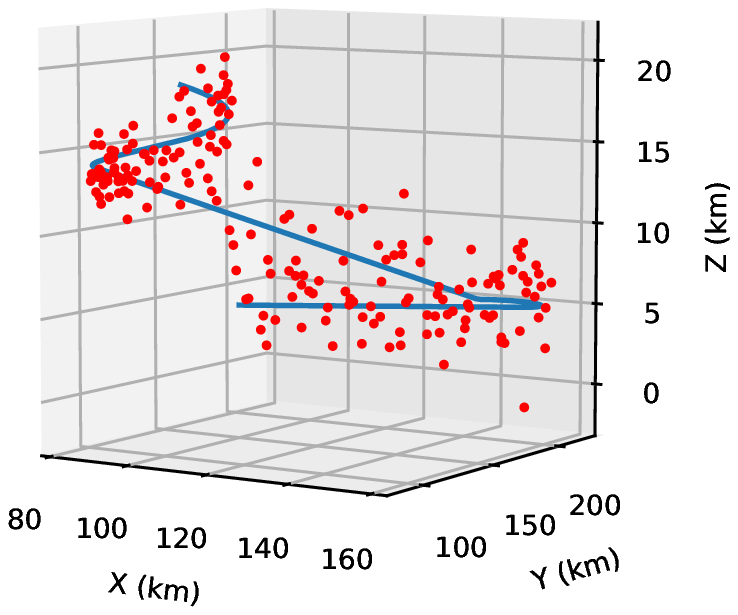}
\caption{Tracks and Measurements}
\label{pred:track}
\end{subfigure}
\begin{subfigure}{.5\textwidth}
  \includegraphics[scale=1]{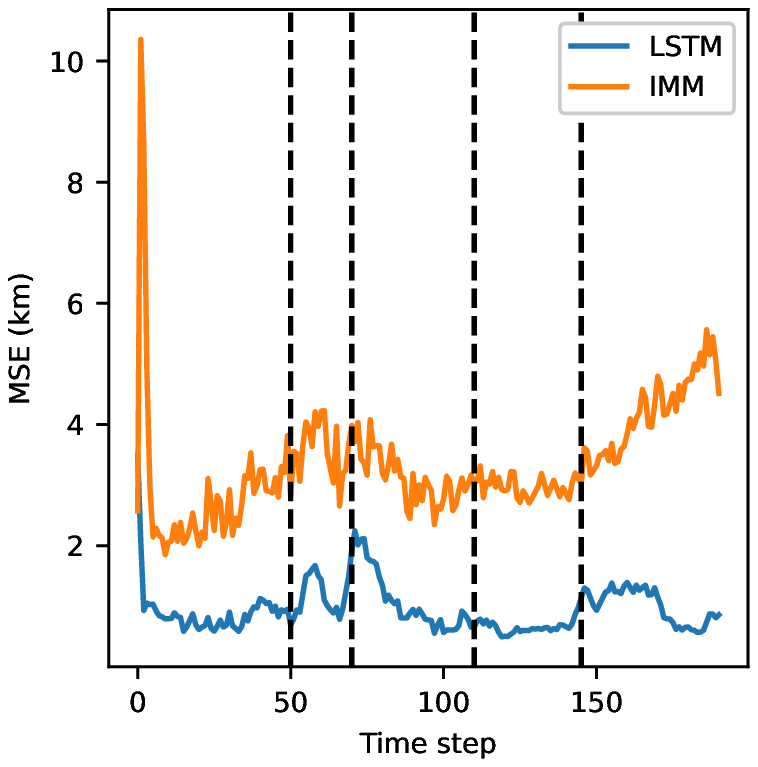}
  \caption{MSE of Position Prediction Model}
  \label{pred:result}
\end{subfigure}
\caption{Simulation}
\end{figure}
\subsubsection{Performance}
\label{pred}
The model is tested on simulated track shown in Fig.~\ref{pred:track} with segments modeled as constant velocity for $50$ time steps, coordinated turn for $20$ time steps, constant linear acceleration of $8m/sec^2$ for $35$ time steps, constant linear acceleration of $-8m/sec^2$ for $35$ time steps and a spiral turn with linear acceleration of $5m/sec^2$ for last $50$ timesteps. The measurement model is same as described in Section \ref{Section:MTT}. The result shown in Fig. ~\ref{pred:result} is an average Mean Squared Error (MSE) between predicted position and ground truth vs time steps of 100 Monte Carlo runs, with vertical lines marking the switches in motion model. It can be seen that the predictions of LSTM model outperforms the IMM model in terms of MSE meric and also from figure it is evident that the LSTM model handles the switches between the motion model considerably better than IMM. The overall MSE of predicted positions of LSTM is $71.64\%$ less than the IMM model.
\subsection{Filter}
\begin{figure}[ht]
 \begin{subfigure}{.5\textwidth}
\centering
\includegraphics[scale=1]{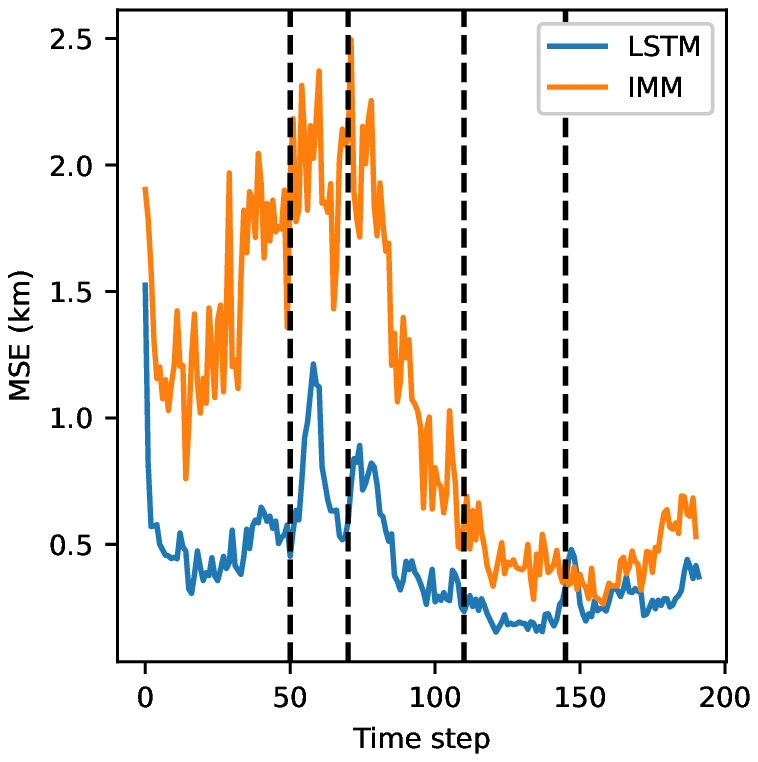}
\caption{MSE of test track}
\label{filter:mse}
\end{subfigure}
\begin{subfigure}{.5\textwidth}
  \includegraphics[scale = 1]{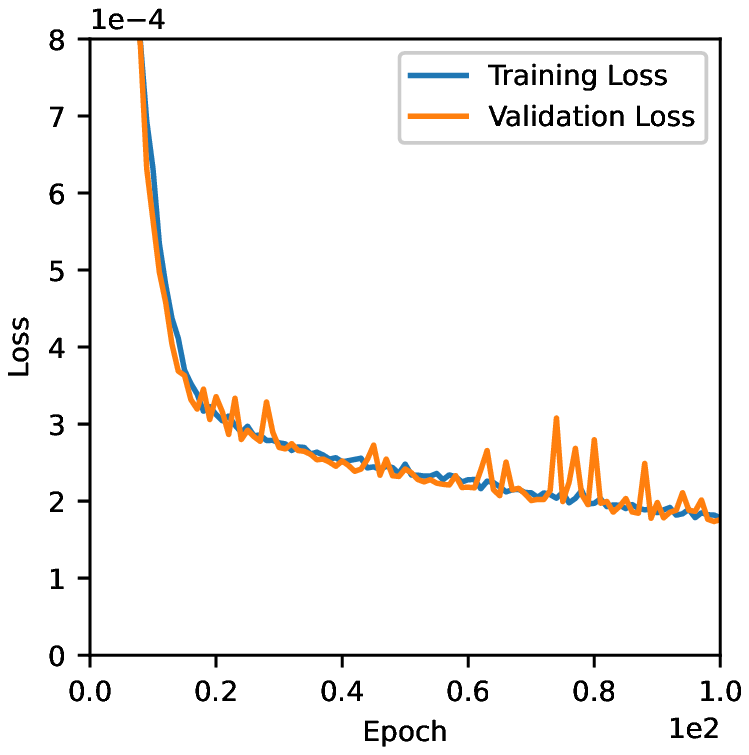}
  \caption{Loss vs Epoch}
  \label{filter:train}
\end{subfigure}
\caption{Filter Model}
\end{figure}
\subsubsection{Model}
In the filter module, a 3 layer LSTM model is used as described in Fig. ~\ref{bilstm}, with hidden layer nodes = $256$ for all layers. Loss function used is Mean Squared Error (MSE), where loss is calculate as Eq.~\eqref{mse}, where $n$ is number of states, $y_{i}$ is ground truth of $i^{th}$ state and $x_{i}$ is predicted value.

\subsubsection{Input}
The input $I^f$ to the model is the associated measurement values, for current time step $t$. $$I^f_k= [x,y,z,v_x,v_y,v_z]_{k}$$ where $I^f_k$is the input for track $k\in[1,m]$ where $m$ is number of current tracks $ x,y,z $ are position coordinates and $ v_x, v_y, v_z $ are components of radial velocity $w.r.t.$ to coordinate axis.

\subsubsection{Training} Model is trained on $40,000$ tracks with sequence length of $100$ each $5$ secs apart with Adam optimizer with learning rate of $0.001$. From Fig.~\ref{filter:train} it can be seen that the network gives good validation loss and is not over-fitting.

\subsubsection{Output}
The output matrix $O^f$ of the model is $O^f_k = [x,y,z]_k$ $k\in[1,m]$ where $m$ is number of current tracks, $ x,y,z $ are filtered position coordinates.
\subsubsection{Performance}
The model is tested on a same simulated track described in Section~\ref{pred}. Fig.~\ref{filter:mse} shows the average MSE between filtered position and ground truth vs time steps of 100 Monte Carlo runs. It can be seen that the LSTM model again outperforms the IMM model in filtering and gives $55\%$ better MSE as compared to the IMM.

\subsection{Track Management}
All unassociated measurements $R$ from the radar goes through an association procedure using gating algorithm in track management module, which maintains a list of probable tracks where the measurement is associated to the track if it falls within the gate of $5$ km. Once the total number of measurements of any probable track exceeds the threshold of 5, a new current track $N$ is formed which from then on wards uses association module for association.

Track management is also responsible to terminate the inactive tracks, if any track is not associated for more than 3 consecutive time steps, it is considered as inactive and is terminated. These numbers are indicative and depend on the target dynamics and sensor characteristic.

\section{Integrated Performance Analysis}
The modules described in the paper can be integrated in different ways, based on sensor under consideration. For the specific case of radar all the modules are required. For co-operative sensor like AIS, ADS-B, IFF etc., association module is not required and can be replaced with a conventional "ID" based association engine. Each module can be tuned separately, and can be even replaced by conventional methods. Such a modular system can be tested independently and enhance the interpretability of the overall algorithm.

\begin{figure}[ht]
  \begin{subfigure}{.5\textwidth}
\centering
\includegraphics[scale = 1]{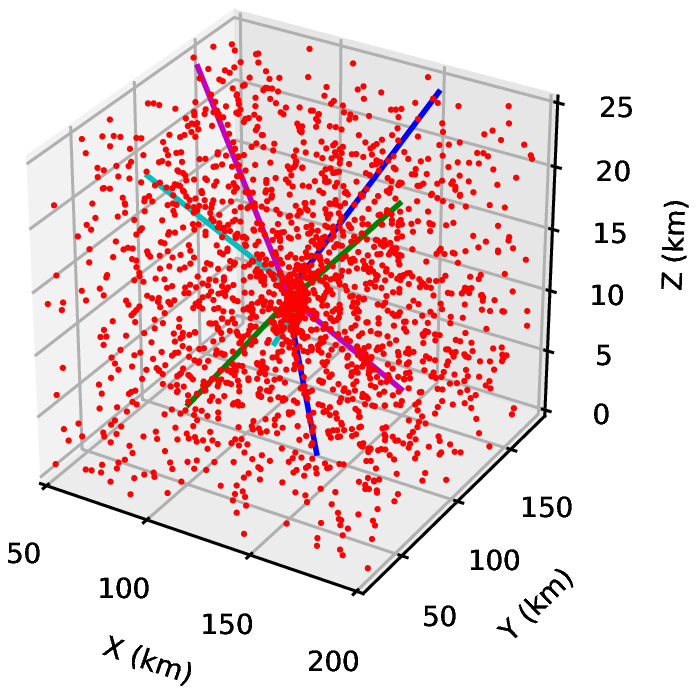}
\caption{Simulated Tracks and Measurements}
\label{multi_tracks}
\end{subfigure}
\begin{subfigure}{.5\textwidth}
  \includegraphics[scale = 1]{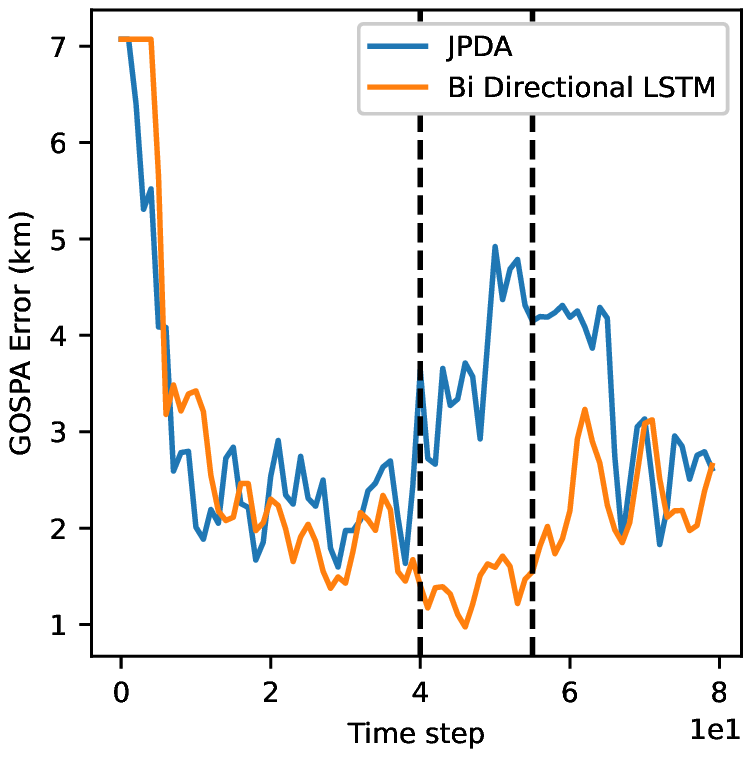}
  \caption{GOSPA metric}
  \label{gospa}
\end{subfigure}
\caption{Integrated Performance}
\end{figure}

We use a radar scenario to illustrate the integrated performance. A crossing path scenario as shown in Fig.~\ref{multi_tracks} is used to show the  effectiveness of the end-to-end model of this paper. The proposed model is compared with the JPDA-IMM filter on Generalized Optimal Sub-Pattern Assignment metric (GOSPA) \cite{Rahmathullah_2017}. The GOSPA distance is calculated using Eq.~\eqref{eq:gospa}, where $X$ and $Y$ denotes the number of established tracks and ground truth receptively, $x$ and $y$ are the estimated position and the ground truth, $c$ is cutoff distance and $p$ is the order the metric, $s$ is switching penalty and $\eta_s$ is number of switches. GOSPA combines localization errors, errors due to missed targets and false targets, and also penalizes for track switching.
\begin{equation}
  \label{eq:gospa}
    GOSPA  = \left(\min\limits_{\pi\in\Pi_{\vert Y\vert}}\sum\limits_{i=1}^{\vert X\vert}d^{(c)}(x_{i},\ y_{\pi(i)})^{p}+\frac{c^{p}}{\alpha}(\vert Y\vert -\vert X\vert)+s\times\eta_s\right)^{\frac{1}{p}} \\
\end{equation}

These are four tracks in the example Fig.~\ref{multi_tracks}. The tracks start with linear velocity of $400$ m/sec each and ascend with constant velocity and pitch angle of $0.17$ radians for $40$ time steps, then all take a coordinated turn with roll angle of $0.26$ radians for $15$ time steps, and finally accelerate with linear acceleration of $10$ $m/sec^2$ for next $40$ time steps. The vertical lines in Fig.~\ref{gospa} denotes the different motion model segments. The measurement model is same as described in the data generation section with $50$ false detections distributed randomly in the space at each time step.

The performance of the models are measured in GOSPA with $p$ = 2, $\alpha$ = 2, $c$ = 7 and $s$ =3. Fig.~\ref{gospa} shows the GOSPA distance of both the models w.r.t. time steps. It can be seen that the Integrated LSTM based model performs better as compared to both JPDA especially when tracks converge to the  intersection point at $40^{th}$ time step, it also maintains the track association accuracy and outperforms the JPDA model. The average GOSPA error of proposed integrated algorithm is $25\%$ less than the JPDA-IMM.
\section{Conclusion}
The design and performance of an end-to-end deep learning based modular blocks for multi-target tracking is elucidated in this paper. The modular design provides robustness and easy adaptability towards different measurement sensors and scenarios. The model free approach handles all the process associated with  the multi target tracking and provides better performance to various target motion models encountered in combat than the conventional approaches. The modular design allows fine tuning of specific modules without having to re-train the whole system. The performance of each individual blocks outperforms the conventional state of the art techniques, which is demonstrated using suitable performance metrics. Also, the end to end GOSPA metric of our technique shows good improvement over conventional JPDA-IMM.

\end{document}